\documentclass{article}

     \PassOptionsToPackage{numbers, compress}{natbib}

\usepackage[final]{neurips_2022}

\usepackage[utf8]{inputenc} 
\usepackage[T1]{fontenc}    
\usepackage{hyperref}       
\usepackage{url}            
\usepackage{booktabs}       
\usepackage{amsfonts}       
\usepackage{nicefrac}       
\usepackage{microtype}      
\usepackage{xcolor}         
\usepackage{bm,bbm,dsfont}
\usepackage{CJK}

\usepackage{color}
\usepackage{tcolorbox}
\usepackage{CJK}
\usepackage{adjustbox}

\usepackage{times}
\usepackage{latexsym}
\usepackage{multirow}
\usepackage{pifont}
\usepackage{amsmath, amsfonts}
\usepackage{algorithm}
\usepackage{algorithmicx}
\usepackage{algpseudocode}
\usepackage{subfigure}

\usepackage{amsmath}
\usepackage{graphicx}
\usepackage{multirow}
\usepackage{multicol}
\usepackage{caption}
\usepackage{subfigure}
\usepackage{enumitem}
\usepackage{float}
\usepackage{makecell}
\usepackage{tabu}

\usepackage[T1]{fontenc}
\usepackage[utf8]{inputenc}
\usepackage{longtable}
\usepackage{color}
\usepackage{balance} 

\definecolor{mygray}{gray}{0.6}

\usepackage{tcolorbox}
\usepackage{cleveref}
\usepackage{bbm}
\usepackage{wrapfig}

\newcommand{\dtrain}{\mathcal{D}_{\text{train}}}

\newcommand{\seedset}{\mathcal{S}_{\text{seed}}}

\newcommand{\ours}{{\textsc{RetroPrompt}}}

\newcommand{\knn}{$k$NN}

\definecolor{right}{RGB}{0,128,96}
\definecolor{wrong}{RGB}{192,0,32}

\definecolor{blue}{rgb}{0,0,0} 

\title{\emph{Decoupling Knowledge from Memorization:} \\ Retrieval-augmented Prompt Learning}

\author{
	Xiang Chen$^{1,2}$\footnotemark[1],
    Lei Li$^{1,2}$\thanks{$\quad$ Equal contribution.} , 
	Ningyu Zhang$^{1,2}$\footnotemark[2] ,
	Xiaozhuan Liang$^{1,2}$, 
	Shumin Deng$^{1,2}$, \\
	\textbf{
	Chuanqi Tan$^{3}$, 
	Fei Huang$^{3}$,
	Luo Si$^{3}$,
	Huajun Chen$^{1,2}$\thanks{$\quad$ Corresponding Author.} 
	} 
	\\
	$^1$Zhejiang University \& AZFT Joint Lab for Knowledge Engine, China \\
	$^2$Hangzhou Innovation Center, Zhejiang University, China \\
	$^3$Alibaba Group, China \\
	\fontsize{11}{10}\selectfont 
	\{xiang\_chen, leili21, zhangningyu, liangxiaozhuan, 231sm, huajunsir\}@zju.edu.cn,  \\
	\fontsize{11}{10}\selectfont \{chuanqi.tcq, f.huang, luo.si\}@alibaba-inc.com \\
}

\begin{document}

\maketitle

\begin{abstract}

Prompt learning approaches have made waves in natural language processing by inducing better few-shot performance while they still follow a parametric-based learning paradigm; the oblivion and rote memorization problems in learning may encounter unstable generalization issues. Specifically, vanilla prompt learning may struggle to utilize atypical instances by rote during fully-supervised training or overfit shallow patterns with low-shot data. To alleviate such limitations, we develop {\ours} with the motivation of decoupling knowledge from memorization to help the model strike a balance between generalization and memorization. In contrast with vanilla prompt learning,  {\ours} constructs an open-book knowledge-store from training instances and implements a retrieval mechanism during the process of input, training and inference, thus equipping the model with the ability to retrieve related contexts from the training corpus as cues for enhancement. Extensive experiments demonstrate that {\ours} can obtain better performance in both few-shot and zero-shot settings. Besides, we further illustrate that our proposed {\ours} can yield better generalization abilities with new datasets. Detailed analysis of memorization indeed reveals {\ours} can reduce the reliance of language models on memorization; thus, improving generalization for downstream tasks\footnote{Code is available in \url{https://github.com/zjunlp/PromptKG/tree/main/research/RetroPrompt}.}.

\end{abstract}

\section{Introduction}

Large parametric language models ~\cite{radfordimproving, Devl2019bert,joshi2020spanbert,bart} have achieved dramatic empirical success in natural language processing (NLP).
Notably, pre-trained language models (PLMs) have learned a substantial amount of in-depth knowledge from data, and have archived tremendous promise in few-shot/zero-shot learning ability with the natural language prompts \cite{gao2020making,DBLP:journals/corr/abs-2110-08207,DBLP:journals/corr/abs-2109-01652}.
However, Recent studies \cite{liu2021gpt,DBLP:journals/corr/abs-2104-08786,DBLP:journals/corr/abs-2203-00902} observe that prompt learning with PLMs usually generalizes unstably in an extremely low-resource setting or emerging domains.
One potential reason is that, it is non-trivial for parametric models to \emph{learn rare or hard patterns well with rote memorization}, thus, resulting in inefficient generalizable performance. 

Intuitively, if we regard the whole training set as a {\it book} and the test phase as the {\it examination}, the current training-test procedure of prompt learning (based on batch data training) can be viewed as {\it page-by-page memorization} and {\it closed-book examination} \cite{meng2021gnnlm}.
During training, vanilla prompt learning may struggle to memorize atypical instances in a fully-supervised setting or overfit shallow patterns with low-shot data \cite{memory,elangovan-etal-2021-memorization}.
Specifically, recent studies\cite{feldman2020does,feldman2020neural} have proposed a long-tail theory, which notes that when the training set has a long-tail distribution and contain small ``sub-populations'' with atypical instances, then PLMs indeed predict on the test data through rote memorizing these atypical instances rather than learning the common patterns \cite{memory,tanzer2022memorisation}.

The limitations of rote memorization remind us of the human learning process of {\emph{``learn by analogy''}} and the proverb that {\emph{``the palest ink is better than the best memory''}}.
Note that humans can perform associative learning to recall relevant skills in deep memories for reinforcing each other, thus, owning the extraordinary abilities to solve few-shot and zero-shot tasks.
Motivated by these, we endeavor to improve the generalization ability of prompt learning with retrieval and association.
Our intuition is that the difficulty of resolving the above limitations can be substantially alleviated if we can decouple the knowledge from memorization by constructing {\it an open-book knowledge-store} from the training data; thus, referring to related knowledge could provide a strong enhancement signal to help the model strike a balance between generalization and memorization.

\begin{wrapfigure}{R}{0.55\textwidth} 
\centering 
\vspace{-0.3cm}
\includegraphics[width=0.55\textwidth]{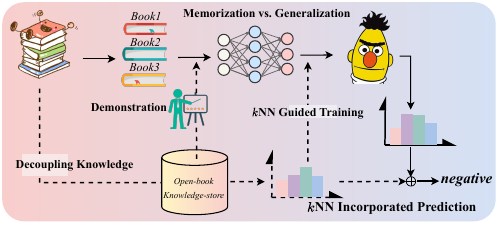} 
\caption{
Decoupling knowledge from memorization.
} 
\label{fig:motivation}
\vspace{-0.2cm}
\end{wrapfigure}

Specifically, we introduce a novel retrieval-augmented framework based on prompt learning (\textbf{\ours}) as shown in Figure~\ref{fig:motivation}.
The open-book knowledge store $\left(\mathcal{K},\mathcal{V}\right)$, defined as the set of \emph{key: prompt-based example embeddings} and \emph{value: corresponding label words} constructed from the training data, are served as additional references for the model to decouple knowledge from pure memorization to some extent.
Specifically, to integrate retrieved knowledge into the input,
\textbf{Firstly}, we design to incorporate neural demonstrations into the input sequences as in-context augmentation, where the demonstration is retrieved from the knowledge-store.
\textbf{Then}, we apply a non-parametric algorithm \knn{} over the input query and knowledge store, and regard \knn{} results as an indication of easy vs. ~hard instances. 
Moreover, we automatically force the model to emphasize the hard instances identified by \knn{} by assigning a scaling during training.
\textbf{Lastly}, the \knn{} results are further employed at the output of the PLM head to participate in masked prediction. 
The model conducts inference through linearly interpolating the non-parametric nearest neighbor distribution with the output of prompt learning, which regards the Top-$k$ nearest reference instances as cues from $\left(\mathcal{K},\mathcal{V}\right)$.
 
The considerable performance gains on nine tasks in few-shot and zero-shot settings demonstrate that our systemic retrieval mechanism helps the model generalize better with scarce data.
Experiments in the fully-supervised setting with long-tail distribution illustrate that our {\ours} can deal with atypical instances more robustly.
We further adopt self-influence \cite{koh2017understanding} as our memorization scoring function to analyze the memorization process between fine-tuning, prompt learning and our {\ours}.
The final analysis results show that 
1) the training samples having the highest memorization scores are mostly atypical,
2) {\ours} generalize better than fine-tuning and convention prompt-tuning with decoupling knowledge from memorization to alleviate the rote of PLMs.
In a nutshell, our work may open up new avenues to improve the generalization of prompting PLMs by decoupling knowledge from memorization.

\section{Preliminaries of Prompt Learning}
\label{background}
Assuming that $\mathcal{M}$, $\mathcal{T}$ respectively denotes the PLM and the template function for prompt tuning. 
Formally, the text classification task takes a query sentence $\bm{x} = (x_0,x_1,...,x_n)$ as input.
Then, classify it into the label ${y} \in \mathcal{Y}$. 
While prompt learning converts the task into a MLM problem with \textit{cloze-style} objectives.
Specifically, the template function $\mathcal{T}$ inserts text pieces into $\bm{x}$ as $\hat{\bm{x}} = \mathcal{T}(\bm{x})$, where $\hat{\bm{x}}$ refers to  the input of $\mathcal{M}$ with a {\tt[MASK]} token.
For instances, when we have to classify the text $\bm{x}$ =``The movie makes absolutely no sense.'' into label \textsc{Negative} (labeled as 0) or \textsc{Positive} (labeled as 1), we wrap it into
\begin{equation}
    \hat{\bm{x}}=
      \texttt{[CLS]} \bm{x} \ \text{It was \texttt{[MASK]}} \texttt{[SEP]}
\end{equation}

The verbalizer $f\colon \mathcal{Y} \mapsto \mathcal{V}$ is defined as a mapping from  the label space $\mathcal{Y}$ to those words in the vocabulary, which constructs the \emph{label word} set $\mathcal{V}$. 
The base component of $\mathcal{M}$ produces the
sequence representation over $\hat{\bm{x}}$, and we choose the hidden vector at the \texttt{[MASK]} position as the contextual  representation $\bm{h}_{\hat{\bm{x}}} \in \mathbb{R}^d$, where $d$ is the dimension of hidden states.
Then the MLM head of $\mathcal{M}$ can operate on $\bm{h}_{\hat{\bm{x}}}$ to calculate each word $v$'s probability in the vocabulary being filled in \texttt{[MASK]} $P_\mathcal{M}(\texttt{[MASK]}=v|\hat{\bm{x}})$.
We let $\mathcal{V}_y$ to represent the subset of $\mathcal{V}$ which is connected with a unique label $y$,  $\cup_{y\in\mathcal{Y}} \mathcal{V}_y = \mathcal{V}$. 
Finally, the probability distribution over the label $y$ is calculated as:
\begin{equation}
\begin{aligned}
   P(y|\bm{x}) \!\!=\!\! g\left(P_\mathcal{M}(\texttt{[MASK]}\!\!\!=\!v|\mathcal{T}(\bm{x}))|v\in\mathcal{V}_y\right),
\end{aligned}
\label{eq:pmscore}
\end{equation}
where $g$ refers to the function converting the probability of label words to the probability of classes.

\section{{\ours}: Retrieval-augmented Prompt Learning}

\begin{figure*}       
    \centering
    \includegraphics[width=1 \textwidth]{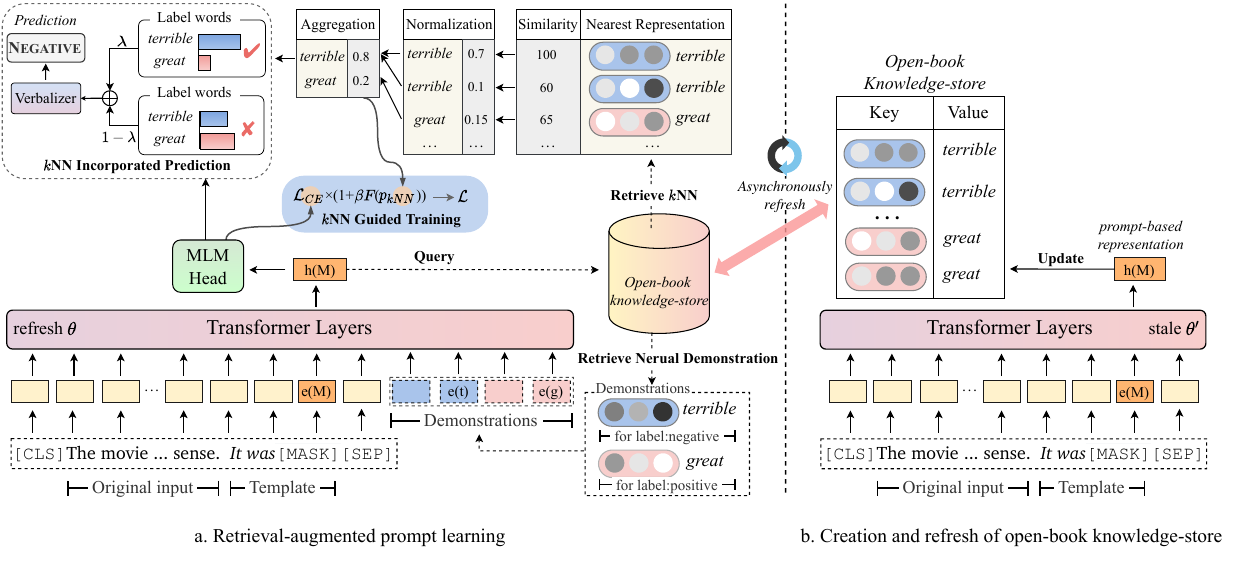}
\caption{Overview of  {\ours}. Note that $e(\cdot)$ denotes word embedding function in the PLM  $\mathcal{M}$,  while ``M'',``t'' and ``g'' in $e(\cdot)$ specifically refers to ``[MASK]'',  ``terrible'' and  ``great''.}
\label{fig:arc}
\end{figure*}

We introduce a simple and general retrieval-augmented framework for prompt learning, named {\ours}, whose basis is the dense retriever (\S \ref{sec:store}) with an open-book knowledge-store to decouple knowledge from memorization.
As shown in Figure \ref{fig:arc},
{\ours} consists of three components: retrieval of neural demonstration for enhancing input  (\S \ref{sec:demo}), the \knn{} guided  training (\S \ref{sec:knn-train}) and the \knn{}-based probability for \textit{cloze-style} prediction (\S \ref{sec:knn-test}).


\subsection{Dense Retriever}
\label{sec:store}

\paragraph{Open-book Knowledge-store}
The first step of our proposed framework is to build a knowledge-store for retrieval that can decouple from memorization and captures the semantics of the instance from the training set  $\mathcal{C}$.
Specifically, we leverage the encoder to embed instance representation over the $\mathcal{C}$ to construct the knowledge-store.
Given the $i$-th example $\left(\bm{c}_i,{y}_i\right)$ in the training data $\mathcal{C}$, 
we obtain the key-value pair $(\bm{h}_{\hat{\bm{c}}_i},v_i)$, in which $\hat{\bm{c}}_i=\mathcal{T}({\bm{c}_i})$,  $\bm{h}_{\hat{\bm{c}}_i} \in \mathbb{R}^d$ is the embedding of the {\tt[MASK]} token in the last layer of the PLM, and $v_{i}=f(y_{i})$ denotes  the label word of the $i$-th example. {
\color{blue}Compared with kNN-LM~\cite{DBLP:conf/iclr/KhandelwalLJZL20} that constructing with sliding generative corpus and tokens, our knowledge-store is more suitable for prompt learning.}
We store all pairs $(\bm{h}_{\hat{\bm{c}}},v)$ in a key-value datastore $\left(\mathcal{K},\mathcal{V}\right) $ where $\bm{h}_{\hat{\bm{c}}}$ serves as \emph{key} and $v$ as \emph{value} as follows:
\begin{equation}
\begin{aligned}
\left(\mathcal{K},\mathcal{V}\right) = 
\{
\left(\bm{h}_{\hat{\bm{c}}{_i}},v_i\right) \mid \left({\bm{c}_i},y_i\right)\in \mathcal{C} 
\}
\end{aligned}
\end{equation}
The knowledge-store maye be flexible to edit, add or delete any instances and can be asynchronously updated during the training procedure. Note that our knowledge-store is constructed from few-shot trainsets in the corresponding few-shot settings rather than the whole available training data.


\paragraph{Efficient Searching}
Considering that the size of the training data $\mathcal{C}$ can be enormous, we must ensure an efficient retrieval process.
As shown in the above creation of open-book knowledge-store, we can build the matrix $\mathbf{D}\in \mathbb{R}^{|\mathcal{C}|\times d}$ as the index of training examples.
Given a query set $Q$, we first encode each query example with template mapping function $\mathcal{T}(\cdot)$ to get a set of prompt-based query vectors $\bm{h}_{\hat{q}}$ for retrieval augmentation on the fly.
Then, we utilize query vectors to search for the closest examples over the index $\mathbf{D}$ via maximum inner product search (MIPS).
For the retrieval process, we choose FAISS~\cite{DBLP:journals/tbd/JohnsonDJ21} to query the open-book knowledge-store efficiently. 
FAISS is an excellent open-sourced toolkit for fast nearest neighbor retrieval. 

\paragraph{Asynchronous Refresh of the Knowledge-store}
\label{refresh}
Since the neural demonstration may lead to the variable contextual representation of instance as the parameters of the PLM are continually updated,
we thus propose to “refresh” the index of retrieval by asynchronously re-embedding and re-indexing all embeddings in an open-book knowledge-store every $j$ training epochs
\footnote{Specifically, we refresh the knowledge-store for each epoch in our experiments.}.
In \S~\ref{ablation}, we empirically demonstrate that this procedure results in performance improvement.

\subsection{Retrieval of Neural Demonstration}
\label{sec:demo}
To enhance the PLMs with the ability to learn by analogy through the knowledge-store, {\color{blue} we further propose neural demonstrations that can be concatenated with input instance at the embedding layer to improve the generalization ability of our {\ours}.}
For the $t$-th query instance $\bm{q}_t$, 
we first utilize prompt-based representation $\bm{h}_{\hat{q}_t}$ to query the cached representations of open-book knowledge-store. 
Then we retrieve $m$ nearest neighbors  $\{ \{\bm{c}^{(1)}_{1}, ..., \bm{c}^{(1)}_{m}\}, ..., \{\bm{c}^{(L)}_{1}, ..., \bm{c}^{(L)}_{m}\}\}$ of $\bm{q}_t$ for each class, where the superscript $L$ denotes the total number of the classes and the $\bm{c}_{i}^{(l)}$ is retrieved as the  $i$-th nearest neighbor in the $l$-th class.
After the model retrieves the Top-$m$ candidates for each class, their corresponding representation $\bm{h}_{\bm{\hat{c}}_{i}}^{(l)}$  and label word $v^{(l)}$ from knowledge-store
will be incorporated into the encoder to act as a demonstration learning. 
Since the $\bm{h}_{\bm{\hat{c}}_{i}}^{(l)}$ is already vector, we intuitively aggregate the $m$  neighbor vectors for each class according to their similarity and 
incorporate the demonstration into the input representation of $\hat{\bm{x}}$ after the word embedding layer of the $\mathcal{M}$ as follows:
\begin{equation}
\small
    \mathcal{I} = {e}(\hat{\bm{x}}) \oplus 
    [\sum_{i \in [1:m]}\alpha_{i}^{(1)} \bm{h}_{\hat{\bm{c}}_i}^{(1)},
    {e}(v^{(1)})] 
    \oplus  ... \oplus 
    [\sum_{i \in [1:m]}\alpha_{i}^{(L)} \bm{h}_{\hat{\bm{c}}_i}^{(L)},
    {e}(v^{(L)})] ;
    \alpha_i^{(l)} = \frac{e^{
    \bm{h}_{\hat{\bm{q}}}
    \cdot \bm{h}_{\hat{\bm{c}}_i}^{(l)}}}
    {\sum_{i \in [1:m]} e^{\bm{h}_{\hat{\bm{q}}}
    \cdot \bm{h}_{\hat{\bm{c}}_i}^{(l)}}}
\end{equation}

where ${e}(\cdot)$ represents the word embedding layer of $\mathcal{M}$, $\oplus$ denotes the concatenation of input sequences, ${\alpha}_{i}^{(l)}$ is the softmax score for the $i$-th retrieval belonging to $l$-th class label to denote their relevance with $\hat{\bm{q}}$, and $\mathcal{I}$ is the sequence features for inputting the next layer of PLM.
As shown in the above equation, we encode demonstration representation with the weighted sum of the retrieval representation. Thus, retrieval scores are directly used in the final representation, making the framework differentiable.
To this end, we denote this style of demonstration as \emph{neural demonstration}, significantly different from prior work of \emph{discrete demonstration}~\cite{gao2020making}.

\textbf{Neural vs. Discrete Demonstration}
Compared with prior discrete demonstrations described in ~\cite{gao2020making,DBLP:journals/corr/abs-2101-06804,DBLP:journals/corr/abs-2112-08633,kumar-talukdar-2021-reordering}, retrieving weighted neural demonstrations from the knowledge-store to augment prompt learning has advantages in the following three major aspects:
(1) neural demonstrations could be more tolerant of the model's maximum input length than discrete demonstrations, while the discrete demonstration is usually not suitable for multi-class classification tasks due to the limitation of input length, such as relation extraction, etc.
(2) the model needs to deal with large retrieval tokens for discrete demonstration, making it time-consuming and computationally intensive to perform cross-attention operations due to the quadratic attention complexity. In contrast,  dealing with much shorter instance representations as neural demonstrations unleashes the potential of cross-attention and accelerates the inference.
(3) when sampling examples based on the similarity between instances, our \textit{cloze-style}  contextual representation is more informative and consistent than the contextual representation from \texttt{[CLS]} of Sentence-BERT~\cite{reimers2019sentence} (adopted in LM-BFF).

\subsection{Retrieve \knn{} for Guiding Training}
\label{sec:knn-train}
Eager learners (e.g., PLMs) are optimized to learn a global function that maps from the text to semantic label space.
Lazy learners such as $k$-nearest neighbor classifiers, on the contrary, aims to approximating the neighborhoods around those test examples~\cite{bontempi2001local}.
Since \knn{} can easily predict for each encountered query instance based on pre-trained representation without an extra classifier, it is intuitively to leverage the \knn{}'s classification results as the \textbf{prior external knowledge} to guide the PLMs' parameters attending to hard examples (hard samples usually refer to atypical samples) during the training process (also referred as \knn{}-train for the abbreviation).
Particularly, our intuition is to differentiate between easy and hard examples according to the prediction of \knn{}. 
Given the $t$-th query instance $\bm{q}_t$, 
we leverage the $\bm{h}_{q_t}$ querying the open-book knowledge-store  $\left(\mathcal{K},\mathcal{V}\right)$ to retrieve the $k$-nearest neighbors $\mathcal{N}$ of $\bm{q}_t$ according to a similarity function $d(\cdot, \cdot)$, where
$d(\cdot, \cdot)$ typically adopt the inner product similarity.
Then, we compute the distribution over neighbors according to the softmax of their similarities and aggregate probability mass for each label word across its occurrences in the retrieved targets:
\begin{equation}
\small
\begin{aligned}
 P_{\text{\knn{}}}\left(y \mid \bm{q}_t \right)  & \propto
 \sum_{\left(\bm{c}_i,y_i\right)\in \mathcal{N}} \mathbbm{1}_{y=y_i} \exp\left(d\left( \bm{h}_{\hat{\bm{q}}_t}, \bm{h}_{\hat{\bm{c}}_i}\right)\right).
\label{eq:knnscore}
\end{aligned}
\end{equation}
Given the probability $p_{k\text{NN}}$ of the query instance $\bm{q}_t$ being predicted as the \textbf{gold class} {\color{blue} (also as the probability value of the gold class in the $P_{k\text{NN}}$)}, we propose to retrieve the \knn{} for guiding the training process of prompt learning. 
The  \knn{} guider reweights the cross-entropy loss $\mathcal{L}_{CE}$ by adjusting the relative loss for the correctly-classified or misclassified instances identified by \knn{}, respectively.
Specifically, we apply the negative log-likelihood as the modulating factor $F(p_{k\text{NN}})$. 
The final loss $\mathcal{L}$ is defined as:
\begin{equation}
\label{eq:joint}
\small
    F(p_{k\text{NN}}) = - \log{}(p_{k\text{NN}}), \quad
    \mathcal{L} = \left(1 + \beta F(p_{k\text{NN}}) \right)\mathcal{L}_{CE},
\end{equation}
where $\beta$ denotes a scalar to determine the proportion of each loss term. 
Note that $p_{k\text{NN}}$ is computed using the \emph{leave-one-out} distribution on the training set due to the fact that each example in the training set cannot retrieve itself.
The motivation of modulating factor is inspired by Focal-loss~\cite{focal_loss}, {\color{blue} while we focus on exploit leveraging k-NN's results for calibrating the training of LMs.}

\subsection{\knn{} based probability for \textit{Cloze-style} Prediction}
\label{sec:knn-test}
Apart from the neural demonstration on the input side and \knn{} guided training process (also referred as \knn{}-test for the abbreviation), we further present \knn{} based probability for \textit{Cloze-style} prediction on the inference process, providing the PLM ability to retrieve nearest neighbors for decisions rather than making predictions only based on memorized parameters.
Given the non-parametric $k$ nearest neighbor distribution $P_{k\text{NN}}$ of the query instance $\bm{q}_t$ being predicted as $y$, we {\color{blue}follow ~\cite{DBLP:conf/nips/GraveCJ17,DBLP:conf/iclr/KhandelwalLJZL20,he2021efficient} to reformulate} the $P(y\mid \bm{q}_t)$ by interpolating the $P_{k\text{NN}}$ with the already-trained base PLM's MLM prediction $P_\mathcal{M}$ using parameter $\lambda$ to produce the final probability of the label:
\begin{equation}
\begin{aligned}
\label{eq:lambda}
P(y \mid \bm{q}_t)=\lambda P_{k\mathrm{NN}}(y \mid \bm{q}_t) 
+(1-\lambda) g\left( P_\mathcal{M} ({\text{\tt [MASK]}} = v|\mathcal{T}(\bm{q}_t)) \right) .
\end{aligned}
\end{equation}
Different from $k$NN-LM~\cite{DBLP:conf/iclr/KhandelwalLJZL20,he2021efficient} that mainly retrieve tokens to augment the language modeling, {\color{blue}we focus on leveraging prompt-based kNN's distribution for reference at test time,  which can unlock the model prediction process as an {\it open-book} examination for prompt learning.}



\section{Experiments}

\begin{table*}[!htp]
\centering
\small
\caption{Results across 9 NLU datasets in the few-shot and zero-shot setting.
We report mean (and standard deviation) results over five different few-shot splits. 
``D-demo'' refers to discrete demonstration, and ``KnPr'' is the abbreviation of KnowPrompt.
LOTClass~\cite{meng2020text} is the SOTA model in unsupervised text classification with self-training. 
{\dag} donates the model uses \textbf{extra knowledge} and {$^\clubsuit$} means they \textbf{train} the PLM on the whole unlabeled trainset, while we and the other baselines only leverage the vanilla  PLM to test without training.
The average scores with {$^*$} denote that we reuse the results of the ``non-demo''  version of the related model to fill in the default values.
}
\scalebox{0.66}{
\begin{tabular}{l|l|lll|lll|l|lll|l}
\toprule
{\multirow{3}{*}{\textbf{St.}}} 
& {\multirow{3}{*}{\textbf{Model}}} 
& \multicolumn{3}{c|}{\textbf{Single Sentence}}
& \multicolumn{3}{c|}{\textbf{Sentence Pair }}
& {\multirow{3}{*}{\textbf{Model}}} 
& \multicolumn{3}{c|}{\textbf{Information Extraction }}
& {\multirow{3}{*}{\textbf{Avg.}}} \\
\cmidrule{3-8}
\cmidrule{10-12}

& &SST-2  & MR   &CR    &MNLI   &QNLI   &QQP  &  &FewN  &SemEval  &TACRED  \\
& & (acc)  & (acc)  & (acc)   & (acc)   & (acc)  & (F1) &   & (acc)   & (acc) & (F1)  &   \\


\midrule
    \multirow{5}{*}{16} 
   & \textsc{FT}~\  
     & 81.4 \tiny{(3.8)}
     & 76.9 \tiny{(5.9)}  
     & 75.8 \tiny{(3.2)} 
     & 45.8 \tiny{(6.4)}
   & 60.2 \tiny{(6.5)} 
   & 60.7 \tiny{(4.3)}
   & \textsc{FT}
   & 52.7 \tiny{(2.2)} 
   & 66.1 \tiny{(1.2)}
   & 25.8 \tiny{(2.8)}
   & 60.6  \\

   & \textsc{LM-BFF} (man)~\  
   & 91.6 \tiny{(1.2 )} 
   & 87.0 \tiny{(2.0)}  
   & 90.3 \tiny{(1.6)} 
   & 64.3 \tiny{(2.5)}
   & 64.6 \tiny{(5.4 )} 
   & 65.4 \tiny{(5.3)} 
   & {KnPr}~\  
   & 65.3 \tiny{(1.1)}  
   & 80.9 \tiny{(2.5)}
   & 33.2 \tiny{(2.0)}
   &  71.4 \\
   
   & \textsc{LM-BFF} (D-demo)
  & 91.8 \tiny{(1.2 )} 
  &  86.6 \tiny{(1.8)}  
  & 90.2 \tiny{(1.4)} 
  & 64.8 \tiny{(2.3)}
  & 69.2 \tiny{(5.4)}
   & 68.2 \tiny{(3.2)}
   & {KnPr} (D-demo)~
   & ~\ ~\  ---
   & ~\ ~\ ---
   & ~\ ~\ ---
   & 72.2{$^*$}  \\
  
    
 & \textsc{KPT} \dag
  & 90.3 \tiny{(1.6)}
  & 86.8 \tiny{(1.8)}  
  & 88.8 \tiny{(3.7)} 
  & 61.4 \tiny{(2.1)} 
   & 61.5 \tiny{(2.8)} 
   & 71.6 \tiny{(2.7)} 
   & \textsc{KPT} \dag 
   &  65.9 \tiny{(1.5)}  
   &  78.8 \tiny{(2.1)}  
   &  32.8 \tiny{(1.7)}
   & 70.9 \\

\cmidrule{2-13}

 & \textbf{Ours}  
 & \textbf{93.9} \tiny{(0.4)} 
 & \textbf{88.0} \tiny{(0.8)} 
 & \textbf{91.9} \tiny{(0.7)}
 & \textbf{71.1} \tiny{(1.8)}
 & \textbf{71.6} \tiny{(1.8)}
 & \textbf{74.0} \tiny{(2.0)}
 & \textbf{Ours}
 & \textbf{67.3} \tiny{(0.9)}  
 & \textbf{81.5} \tiny{(1.3)}  
 & \textbf{40.7} \tiny{(0.7)}  
 & \textbf{75.6} \\
 
 \midrule
    \multirow{5}{*}{4} 
   & \textsc{FT}~\  
   & 60.2 \tiny{(2.8)} 
    & 57.6 \tiny{(1.4)}
   & 66.4 \tiny{(5.5)}
   & 35.0 \tiny{(0.3)}
   & 54.2 \tiny{(3.9)} 
   & 52.8 \tiny{(4.7)}  
   & \textsc{FT} 
   & 32.7 \tiny{(2.9)} 
   & 38.8 \tiny{(2.0)}
   & 14.7 \tiny{(2.8)}
   & 45.8 \\
   
   & \textsc{LM-BFF} (man)~\  
   & 90.7 \tiny{(0.8)} 
   & 85.2 \tiny{(2.8)}
   & 89.9 \tiny{(1.8)} 
   & 51.0 \tiny{(2.5)}
   & 61.1 \tiny{(6.1)} 
   & 48.0 \tiny{(4.9)} 
   & {KnPr}
   &  52.5 \tiny{(1.5)}  
   &  58.4 \tiny{(3.7)}
   &  28.8 \tiny{(2.5)}
   &  62.8\\
   
   & \textsc{LM-BFF} (D-demo)~\  
   & 90.2 \tiny{(1.5)}
   & 85.5 \tiny{(2.1)}
   & 89.7 \tiny{(0.6)} 
   & 56.1 \tiny{(1.0)}
   & 61.7 \tiny{(7.6)} 
   & 63.2 \tiny{(5.6)} 
   & {KnPr} (D-demo)
   &  ~\  ---
   &  ~\  ---
   &   ~\  ---  
   &  {65.1}{$^*$}  \\

   & \textsc{KPT} \dag
   &  88.2 \tiny{(5.7)} 
   &  83.4 \tiny{(1.5)}  
   &  87.2 \tiny{(2.5)}
   &  53.7 \tiny{(2.7)}
   &  59.2 \tiny{(2.8)} 
   & 54.9  \tiny{(7.9)}
   & \textsc{KPT} \dag
   & 58.8 \tiny{(2.2)}   
   & 57.2 \tiny{(3.2)}
   & 27.5 \tiny{(2.2)}
   & 63.3 \\

\cmidrule{2-13}
     & \textbf{Ours}  
      &  \textbf{91.5} \tiny{(1.8)}
      &  \textbf{87.4} \tiny{(0.5)}
      &  \textbf{91.4} \tiny{(0.6)} 
      &  \textbf{57.6} \tiny{(5.5)}
      &  \textbf{62.2} \tiny{(6.0)}
      &  \textbf{66.1} \tiny{(4.1)}
      & \textbf{Ours}
      &  \textbf{60.9} \tiny{(1.9)} 
      & \textbf{59.2} \tiny{(3.0)} 
      & \textbf{32.1} \tiny{(2.0)}
      & \textbf{67.6} \\
 
\midrule
    \multirow{7}{*}{0} 

   & {LOTClass}$^\clubsuit$ ~\  
   &  71.8
   &  81.7
   &  50.1
   &  50.4
   &  36.5
   & 55.9
   & {LOTClass}$^\clubsuit$ ~\ 
   & 11.5
   & 9.8
   & 2.5
   &  41.1 \\
   
   & \textsc{FT}~\  
   &  49.1
   &  50.0
   &  49.8
   &  34.4
   &  49.5
   &  31.6
   & {FT}
   & 10.0
   & 6.2
   & 0.5
   & 31.2 \\
   
   & \textsc{LM-BFF} (man)~\  
   & 83.5  
   & 80.3
   & 78.4
   & 49.7
   & 50.5
   & 49.7
   & {KnPr} 
   & 15.9 
   & 10.3
   & 2.3
   & 46.7 \\
   
     & \textsc{LM-BFF} (D-demo)~\  
     & 82.9 
     & 80.7
     & \textbf{81.4}
     & 52.2
     & 53.5
     & 44.0
     & {KnPr} (D-demo)
     &  ~\  ---
     &  ~\  ---
     &  ~\  ---  
     & 47.0{$^*$} \\
    
   & \textsc{KPT} \dag
   &  78.4
   &  81.9
   & 71.4
   &  37.1
   &  55.3
   &  47.5
   & \textsc{KPT} \dag
   &  24.6
   & 11.6
   &  0.8
   & 45.7 \\

\cmidrule{2-13}
      & \textbf{Ours} 
      &  \textbf{86.8}
      &  \textbf{83.5}
      &  {79.7}
      &  \textbf{53.7}
      &  \textbf{56.2}
      &  \textbf{56.7}
     & \textbf{Ours} 
      &  \textbf{41.3}
      & \textbf{12.2}
      & \textbf{2.8}
      &  \textbf{52.5} \\
\bottomrule

\end{tabular}
}
\label{tab:experiment-few-shot}
\end{table*}

\subsection{Datasets and Baselines }

\textbf{Datasets}
We evaluate {\ours} on several types of natural language understanding tasks, including single sentence classification tasks (SST-2~\cite{sst2}, MR~\cite{mr}, and CR~\cite{cr}) and sentence pair classification tasks (MNLI~\cite{mnli}, QNLI~\cite{qnli}, and QQP\footnote{\url{https://www.quora.com/q/quoradata/}.}). 
To further evaluate the effectiveness of the proposed approach with multi-class classification, we also conduct experiments on the information extraction tasks,
including FewNERD~\cite{fewnerd}, SemEval 2010 Task 8 (SemEval)~\cite{hendrickx2010semeval}, and TACRED~\cite{DBLP:conf/emnlp/ZhangZCAM17}.
The detailed statistics of the datasets are shown in Appendix A. 

\textbf{Baselines}
We compare with LM-BFF~\cite{gao2020making}  for single sentence and sentence pair classification tasks and adopt SOTA prompt learning model KnowPrompt~\cite{chen21knowprompt} as the baseline for information extraction tasks.
Note that the discrete demonstration method cannot be applied to multi-class classification tasks due to the input length limitations; thus, we leave out the experimental table about the results of KnPr (D-demo).
We also compare our {\ours} with the knowledge-enhanced prompt learning method KPT~\cite{KPT} since KPT leverages the external knowledge base for enhancing prompt learning while we focus on utilizing internal trainsets as a knowledge-store.
You can refer to Appendix B for the detailed introduction of baseline methods.

\subsection{Evaluation protocols and details}
\label{subsec:details}

The experiments are implemented on 1 NVIDIA V100 and utilize Pytorch \cite{DBLP:conf/nips/PaszkeGMLBCKLGA19} as the base library. 
We adopt $\text{RoBERTa}_\text{large}$~\cite{liu2019roberta} as the PLM and employ AdamW as the optimizer for all experiments.
To mitigate the influence of diverse templates, we conduct baselines and {\ours} with the same templates for each dataset.
We list the specific experimental settings and tuning retrieve parameters in Appendix C and D.
As for few-shot and zero-shot experiments, we leverage different settings, respectively.

\begin{wrapfigure}{L}{0.3\textwidth} 
\centering 
\includegraphics[width=0.3\textwidth]{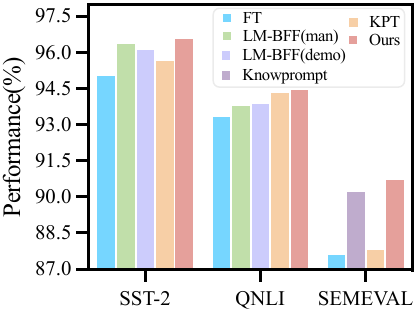} 
\caption{Performance on fully-supervised datasets.}
\label{fig:fully_supervised}
\end{wrapfigure}

\textbf{Few-shot Setting.}
We follow the few-shot setting of LM-BFF \cite{gao2020making} to conduct 4-shot and 16-shot experiments and
evaluate the average performance with a fixed set of seeds, $\seedset$, across several different sampled $\dtrain$ for each task.
Note that our knowledge-store is constructed with the \textbf{few-shot training set} in this setting.

\textbf{Zero-shot Setting\footnote{\color{blue}Note that it is not a strict zero-shot sense.}.}
We leverage vanilla $\text{RoBERTa}_\text{large}$ for all baselines (except LOTClass~\cite{meng2020text}) to directly inference on the test set.
To take advantage of retrieval mechanism, {\ours} follows LOTClass~\cite{meng2020text} to utilize \textbf{unlabeled} trainsets for retrieval.
Specifically, we take the vanilla $\text{RoBERTa}_\text{large}$ to tag the pseudo labels on unlabeled trainset and create the open-book knowledge-store with the unlabeled trainsets and pseudo labels.
Lastly, {\ours} make predictions on the test set based on the constructed datastore \textbf{without tuning any of the model parameters}.



\subsection{Experimental Results}


\textbf{Few-shot Results.}\quad 
As shown in Table~\ref{tab:experiment-few-shot}, we find {\ours} consistently outperforms baseline method LM-BFF and KnowPrompt, both in 4-shot and 16-shot experiments. 
Especially for information extraction tasks with multiple classes, discrete demonstrations cannot be applied to the input due to the limited input sequence length, while our neural demonstration can also work and achieves improvement on these multi-class datasets.
Moreover, {\ours} obtain better performance compared with KPT. Compared with KPT with external knowledge, we only focus on referencing the internal few-shot trainsets without visiting the external knowledge base. 
Besides, we observe that {\ours} has a relatively lower standard deviation than the baselines. 
The reason may lie that the retrieval mechanism can compensate for instabilities in parametric predictions.



\begin{wraptable}{r}{0.45\textwidth}
\centering
\small
\vspace{-0.2cm}
\caption{Results of model generalization to new domains.}
\scalebox{0.7}{
\begin{tabular}{l|c|cc}
\toprule
 {\multirow{1}{*}{\textbf{Model}}} 

& \multicolumn{1}{c|}{\textbf{Source}}
& \multicolumn{2}{c}{\textbf{Target Domain}}
\\
\cmidrule{1-4}
&16-shot MR &SST-2 & CR   \\

\midrule

    \textsc{FT}~\  
   & 76.9
   & 71.4
   & 64.7

  \\
   
    \textsc{LM-BFF} (man)~\  
   & 87.0
   & 88.9
   & 86.9
 \\
   
    \textsc{LM-BFF} (D-demo)~\  
     & 86.6 & 89.3  & 87.5
\\
    
  \textsc{KPT} 
   & 86.8 & 86.8  & 86.7
\\

\cmidrule{1-4}
     \textbf{\ours}  
      &  \textbf{88.0}  &   \textbf{91.4}  &   \textbf{88.8}
\\
\cmidrule{1-4}
 &16-shot QQP &MRPC & RTE   \\

\midrule
    \textsc{FT}~\  
   & 60.7
   & 43.7
   & 48.0

  \\
   
    \textsc{LM-BFF} (man)~\  
   & 65.4
   & 20.9
   & 65.5
 \\
   
    \textsc{LM-BFF} (D-demo)~\  
     & 68.2
     & 38.8
     & 66.2
\\
    
  \textsc{KPT} 
   & 71.6
   & 42.3
   & 65.8
\\

\cmidrule{1-4}
     \textbf{\ours}  
      &  \textbf{74.0}  
      & \textbf{49.4}  &   \textbf{67.3}
 \\
\bottomrule
\end{tabular}
}
\label{tab:experiment-cross-domain}
\end{wraptable}

\textbf{Zero-shot Results.}\quad
From Table~\ref{tab:experiment-few-shot}, we also observe that {\ours} achieves improvements in the zero-shot setting.
Another notable point is that
{\ours} performs even better than KPT in the zero-shot setting, revealing that exploring own data to decouple knowledge from memorization has more potential than leveraging external knowledge.
Moreover, we achieve superior performance to LOTClass even though we utilize the vanilla $\text{RoBERTa}_\text{large}$ without any training.

\textbf{Fully-supervised Results.}\quad
As shown in Figure~\ref{fig:fully_supervised},
the experiments in fully-supervised settings with long-tail distribution illustrate that {\ours} achieves improvement compared with baselines. 
This indicates that our retrieval mechanism extends the LM's ability to learn hard examples in the fully-supervised datasets.

\subsection{Model Generalization to New Domains}
The scarce data may bring the overfitting problem for the lots of memory parameters of PLMs, even though prompt learning.
Thus, we conduct cross-domain experiments to validate the generalization of our {\ours}. 
Specifically, we utilize the model trained on the source datasets and directly test on the other target datasets.
From Table~\ref{tab:experiment-cross-domain}, we can find that our method  consistently outperforms baselines. This finding illustrates that {\ours} achieves great model generalization to new domains.

\subsection{Analysis of Memorization}
\label{subsec:memorization}
It is necessary and interesting to further explore the memorization mechanism to help us better understand the utility of retrieval for memorization in NLP.


\textbf{Definition of Memorization Measurement.}\quad
Inspired by the idea of \cite{feldman2020does} in the computer vision area, we define {\it memorization measures} as to how the classification varies when a training instance $\bm{z}$ is deleted from the trainset.
We follow \cite{koh2017understanding,memory} to define and derive the memorization score for a training instance $\bm{z}$ as follows:
\begin{equation}
\small
\label{equ:remove}
\begin{aligned}
    {\mathcal{S}}_{\text{delate}}(\bm{z}) 
    &\overset{\text{def}}{=} -\frac{d P(y|\bm{x}; \Hat{\theta}_{\xi, -\bm{z}})}{d \xi} \bigg|_{\xi=0}
    &= -\nabla_{\theta}P(y|\bm{x}; \hat{\theta})^{\top}\frac{d \hat{\theta}_{\xi, -\bm{z}}}{d \xi} \bigg|_{\xi=0} 
    &= -\nabla_{\theta}P(y|\bm{x}; \hat{\theta})^{\top}H^{-1}_{\hat{\theta}}\nabla_{\theta}{\mathcal{L}(\bm{z}, \hat{\theta})},
\end{aligned}
\end{equation}
where $\hat{\theta}_{\xi, -\bm{z}}$ denotes the parameters trained with the instance $\bm{z}$ down-weighted by $\xi$, $\hat{\theta}$ refers to the parameters of the model trained with all instances and $H_{\hat{\theta}} = \frac{1}{n}\sum^{n}_{i=1}{\nabla^{2}_{\theta}{\mathcal{L}(z_i, \hat{\theta})}}$.
Thus $\mathcal{S}_{\text{delate}}(\bm{z})$ refers to the amount of change of $P(y|x; \theta)$ when the instance $\bm{z}$ is down-weighted by $\xi$.

\begin{table*}[!htp]
\centering
\small
\caption{The upper part shows the average percentage of \emph{positive phrases} over different memory groups of positive/negative instances. The lower part denotes the mean values of memorization score on the SST-2 dataset.
}
\label{table.atypical}
\begin{small}
\scalebox{0.8}{

\begin{tabular}{l|ccc|ccc}
\toprule
\multirow{2}{*}{\textbf{Mem Group}} & \multicolumn{3}{c}{\textbf{Negative}} & \multicolumn{3}{c}{\textbf{Postive}} \\

\cmidrule{2-7}
&FT  & LM-BFF & OURS 
&FT  & LM-BFF & OURS \\

\midrule
Top-10\%  &34.29  & 32.78  & 30.23
          & 68.75  & 69.71 &75.67 \\
ALL                        & \multicolumn{3}{c}{23.40}      & \multicolumn{3}{c}{86.39}        \\

Bottom-10\% & 17.63 & 16.25 & 14.42
& 95.92 & 95.08 & 94.53 
\\

\bottomrule
\midrule

 & \multicolumn{2}{c}{FT} & \multicolumn{2}{c}{LM-BFF} & \multicolumn{2}{c}{OURS} \\
\midrule

\textsc{Mem Score}            & \multicolumn{2}{c}{4.597}   & \multicolumn{2}{c}{0.121}       & \multicolumn{2}{c}{0.032}  \\
\bottomrule
\end{tabular}
}
\end{small}
\end{table*}

\textbf{Top-memorized Instances: Typical or Atypical?}\quad
Since the SST-2 dataset provides the annotations of phrase-level sentiment polarity labels, we adopt SST-2 to analyze the memorization by judging the atypical of an instance by checking the percentage of positive phrases.
We achieve such statistics from SST-2 and observe that a typical positive instance has a relatively high percentage of positive phrases, and a typical negative instance should have a relatively low percentage of positive phrases.
Based on the above observation, we apply the memorization score defined in Eq.~\ref{equ:remove} to select Top-10\% and Bottom-10\% memorized instances from the trainset and collect the average percentage of positive phrases in these instances.

As shown in Table~\ref{table.atypical},we can conclude following findings:
(1) \textbf{The PLM tends to give atypical samples deeper memory attention.}
Specifically, no matter LM-BFF or our method, the top-10\% memorized negative instances have a higher percentage of positive phrases than the average percentage of positive phrases of all negative instances.
2) LM-BFF has lower memorization scores on hard samples than fine-tuning. We think it owns to \textbf{prompt learning can help PLMs recall what they learned from pre-training without strengthening memory for downstream data.} 3) {\ours} further has lower average memorization scores than fine-tuning and LM-BFF, which illustrates that our method is less memory dependent. This result may be attributed to  \textbf{decoupling knowledge from memorization through retrieval to alleviating the rote of PLMs.}

\begin{wraptable}{r}{0.45\textwidth}
\centering
\caption{Detailed ablation experiments in few-shot settings. 
``N-demo'' donates the neural demonstration, 
and ``refresh'' refers to the asynchronous refresh of the knowledge-tore.
}
\scalebox{0.69}{
\begin{tabular}{l|ccccc}
\toprule
\multirow{2}{*}{\textbf{Model}}                      
& \multicolumn{5}{c}{\textbf{16-shot}} 
\\ 
\cmidrule{2-6} 
  & SST-2  & CR & MNLI & QQP & TACRED  
  \\ \midrule
\textbf{OURS}
& \textbf{93.9} & \textbf{91.9} 
 & \textbf{71.1}
& \textbf{74.0}    & \textbf{40.7}  
    \\

\midrule
w/o \text{\knn{}}-test
& 93.2 & 91.2
& 70.4
&  73.0  & 38.2    \\
w/o \text{\knn{}}-train
& 92.0
& 90.2
& 68.8
&  71.3  & 36.5     \\
w/o N-demo
& 92.4
& 91.0 
& 70.1
&  72.7 & 37.9      \\ 
w/o refresh
& 93.5
& 91.5 
& 70.7
&  73.6   & 39.9   \\ 
\bottomrule
\end{tabular}}
\label{tab:ablation}
\end{wraptable}

\textbf{Case Analysis.}\quad
As shown in Table~\ref{table.vis.sst},
we manually list the bottom-ranked and top-ranked training instances of SST-2 according to our model. 
It reveals that the top-ranked memorized instances seem to show universal opinions indirectly. 
Thus, we inspect them as atypical/hard for sentiment classification.
While those instances with 0 memorization scores are straightforward to show their opinion for sentiment classification, representing the typical instance. Note that {$F(p_{kNN})$} is defined to represent the difficulty of the sample discriminated by \knn{} distribution. And the Table~\ref{table.vis.sst} also shows that {$F(p_{kNN})$} indeed reflect atypicality of examples, which validates the effectiveness of the \knn{} guided training.


\subsection{Ablation Study}
\label{ablation}
\paragraph{Component Ablation.}\quad 
As shown in Table~\ref{tab:ablation}, the performance of component ablation experiments with four variants has a clear drop, which validates the power of our retrieval component. 
We also find that neural demonstration and \knn{}-train have more improvement in the few-shot setting than \knn{}-test. 
Note that \knn{}-test is similar to \knn{}-LM~\cite{DBLP:conf/iclr/KhandelwalLJZL20,he2021efficient} and the results reveals that
simply incorporate \knn{} in the test process of prompt learning has little influence in a few-shot setting.

\begin{wraptable}{r}{0.38\textwidth}
    \caption{Performance on 16-shot CR and TACRED with different representations of key and calculate function of \knn{} distribution.}
    \label{result-ktype}
    \centering
    \scalebox{0.75}{
    \begin{tabular}{llcc}
    \toprule
    Key Repres. &  \knn{} Acq. & CR & TAC.  \\
    \midrule
    Prompt & Rep-similar & 91.9 & 40.7\\
    \texttt{[CLS]}  & Rep-similar & 89.0 & 37.2    \\      
    Prompt  & BM25 & 89.5 & 38.8 \\
    \texttt{[CLS]} & BM25 & 88.7 & 36.1 \\ 
    \bottomrule
    \end{tabular}
    }
\end{wraptable}

\paragraph{Key Representation and \knn{} Acquisition.}\quad  We study the effect of using different representations of the key in the knowledge-store. We experiment with two types of representations: (1) prompt-based representation, which is the default setting, and (2) [CLS] based representation of current LM. We also experiment with two types of calculation of \knn{} distribution: (1) representation based similarity score (refer as rep-similar), which is the default setting, and (2) BM25 based score , which calculates the correlation score between the query and each key examples with BM25~\cite{bm25} algorithm.
Results in Table~\ref{result-ktype} show that using prompt-based representations for key and representation based similarity scores for \knn{} leads to the best performance. It suggests that prompt learn better representations for context similarity and the representation similarity based \knn{} distribution is better than BM25 based scores.

\begin{table*}
\centering
\caption{Case examples of Top-3 and Bottom-3 memorized instance of {ours} from trainset of SST-2.
}
\label{table.vis.sst}
\resizebox{\textwidth}{!}{
\begin{tabular}{lp{0.5cm}p{1.3cm}<{\centering}lp{0.5cm}p{1.3cm}<{\centering}}
\toprule
\multicolumn{3}{c}{\textbf{Negative}} & \multicolumn{3}{c}{\textbf{Positive}} \\
\cmidrule(lr){1-3} \cmidrule(lr){4-6}
\textbf{Content} &\textbf{Mem} & {$F(p_{k\text{NN}})$} & \textbf{Content} &\textbf{Mem} & $F(p_{k\text{NN}})$\\
\midrule
\begin{CJK*}{UTF8}{gbsn}
{\setlength{\fboxsep}{0pt}
\colorbox{white!0}{\parbox{0.45\textwidth}{
\colorbox{red!0}{\strut Although} \colorbox{red!0}{\strut god} \colorbox{red!0}{\strut is} \colorbox{red!0}{\strut great} \colorbox{red!0}{\strut addressed} \colorbox{red!0}{\strut interesting} \colorbox{red!0}{\strut matters} \colorbox{red!0}{\strut of} \colorbox{red!0}{\strut identity} \colorbox{red!0}{\strut and} \colorbox{red!0}{\strut heritage,} \colorbox{red!0}{\strut it's} \colorbox{red!0}{\strut hard} \colorbox{red!0}{\strut to} \colorbox{red!0}{\strut shake} \colorbox{red!0}{\strut the} \colorbox{red!0}{\strut feeling} \colorbox{red!0}{\strut that} \colorbox{red!0}{\strut it} \colorbox{red!0}{\strut was} \colorbox{red!0}{\strut intend} \colorbox{red!0}{\strut to} 
\colorbox{red!0}{\strut be} 
\colorbox{red!0}{\strut a} 
\colorbox{red!0}{\strut different} 
\colorbox{red!0}{\strut kind} 
\colorbox{red!0}{\strut of} 
\colorbox{red!0}{\strut film.} 
}}}
\end{CJK*} 
& 0.066 & 1.17 &
\begin{CJK*}{UTF8}{gbsn}
{\setlength{\fboxsep}{0pt}\colorbox{white!0}{\parbox{0.45\textwidth}{
\colorbox{red!0}{\strut A} \colorbox{red!0}{\strut b-movie} \colorbox{red!0}{\strut you} \colorbox{red!0}{\strut can} \colorbox{red!0}{\strut sit} \colorbox{red!0}{\strut through,} \colorbox{red!0}{\strut enjoy} \colorbox{red!0}{\strut on} \colorbox{red!0}{\strut a} \colorbox{red!0}{\strut certain} \colorbox{red!0}{\strut level} \colorbox{red!0}{\strut and} \colorbox{red!0}{\strut then} \colorbox{red!0}{\strut forget.}
}}}\end{CJK*} 
& 0.020 & 0.18\\ 
\begin{CJK*}{UTF8}{gbsn}
{\setlength{\fboxsep}{0pt}\colorbox{white!0}{\parbox{0.45\textwidth}{
\colorbox{red!0}{\strut A} \colorbox{red!0}{\strut standard} \colorbox{red!0}{\strut police-oriented} \colorbox{red!0}{\strut drama} \colorbox{red!0}{\strut that,} \colorbox{red!0}{\strut were} \colorbox{red!0}{\strut it} \colorbox{red!0}{\strut not} \colorbox{red!0}{\strut for} \colorbox{red!0}{\strut deniro's} \colorbox{red!0}{\strut participation,} \colorbox{red!0}{\strut would} \colorbox{red!0}{\strut have} \colorbox{red!0}{\strut likely} \colorbox{red!0}{\strut wound} \colorbox{red!0}{\strut up} \colorbox{red!0}{\strut a} \colorbox{red!0}{\strut tnt} \colorbox{red!0}{\strut original.} 
}}}
\end{CJK*} 
& 0.011 & 1.48 &
\begin{CJK*}{UTF8}{gbsn}
{\setlength{\fboxsep}{0pt}\colorbox{white!0}{\parbox{0.45\textwidth}{
\colorbox{red!0}{\strut A} \colorbox{red!0}{\strut film} \colorbox{red!0}{\strut that} \colorbox{red!0}{\strut will} \colorbox{red!0}{\strut be} \colorbox{red!0}{\strut best} \colorbox{red!0}{\strut appreciated} \colorbox{red!0}{\strut by} \colorbox{red!0}{\strut those} \colorbox{red!0}{\strut willing} \colorbox{red!0}{\strut to} \colorbox{red!0}{\strut endure} \colorbox{red!0}{\strut its} \colorbox{red!0}{\strut extremely} \colorbox{red!0}{\strut languorous} 
\colorbox{red!0}{\strut rhythms,} 
\colorbox{red!0}{\strut waiting} 
\colorbox{red!0}{\strut for} 
\colorbox{red!0}{\strut happiness} 
\colorbox{red!0}{\strut is} 
\colorbox{red!0}{\strut ultimately} 
\colorbox{red!0}{\strut thoughtful} 
\colorbox{red!0}{\strut without} 
\colorbox{red!0}{\strut having} 
\colorbox{red!0}{\strut much} 
\colorbox{red!0}{\strut dramatic} 
\colorbox{red!0}{\strut impact.} 
}}}\end{CJK*}
& 0.010 & 0.43 \\ 
\begin{CJK*}{UTF8}{gbsn}
{\setlength{\fboxsep}{0pt}\colorbox{white!0}{\parbox{0.45\textwidth}{
\colorbox{red!0}{\strut A} \colorbox{red!0}{\strut hit} \colorbox{red!0}{\strut and} \colorbox{red!0}{\strut miss} \colorbox{red!0}{\strut affair,} \colorbox{red!0}{\strut consistently} \colorbox{red!0}{\strut amusing} \colorbox{red!0}{\strut but} \colorbox{red!0}{\strut not} \colorbox{red!0}{\strut as} \colorbox{red!0}{\strut outrageous} \colorbox{red!0}{\strut or} \colorbox{red!0}{\strut funny} \colorbox{red!0}{\strut as} \colorbox{red!0}{\strut cho} \colorbox{red!0}{\strut may} \colorbox{red!0}{\strut have} \colorbox{red!0}{\strut intended} \colorbox{red!0}{\strut or} \colorbox{red!0}{\strut as} \colorbox{red!0}{\strut imaginative} \colorbox{red!0}{\strut as} 
\colorbox{red!0}{\strut one} 
\colorbox{red!0}{\strut might} 
\colorbox{red!0}{\strut have} 
\colorbox{red!0}{\strut hoped.} 
 
}}}
\end{CJK*} 
& 0.010 &  2.74 &
\begin{CJK*}{UTF8}{gbsn}
{\setlength{\fboxsep}{0pt}\colorbox{white!0}{\parbox{0.45\textwidth}{
\colorbox{red!0}{\strut What's} \colorbox{red!0}{\strut invigorating} \colorbox{red!0}{\strut about} \colorbox{red!0}{\strut is} \colorbox{red!0}{\strut that} \colorbox{red!0}{\strut it} 
\colorbox{red!0}{\strut doesn't} 
\colorbox{red!0}{\strut give} 
\colorbox{red!0}{\strut a} 
\colorbox{red!0}{\strut damn.} 
}}}\end{CJK*}
& 0.003 & 0.06 \\ 
\midrule
\begin{CJK*}{UTF8}{gbsn}
{\setlength{\fboxsep}{0pt}\colorbox{white!0}{\parbox{0.45\textwidth}{
\colorbox{red!0}{\strut It's} \colorbox{red!0}{\strut a} \colorbox{red!0}{\strut loathsome} \colorbox{red!0}{\strut movie,} \colorbox{red!0}{\strut it}  
\colorbox{red!0}{\strut really}  
\colorbox{red!0}{\strut is}  
\colorbox{red!0}{\strut and}  
\colorbox{red!0}{\strut it}  
\colorbox{red!0}{\strut makes}  
\colorbox{red!0}{\strut absolutely}  
\colorbox{red!0}{\strut no}  
\colorbox{red!0}{\strut sense.}  
}}}\end{CJK*} 
& 0.00 & 0.00 &
\begin{CJK*}{UTF8}{gbsn}
{\setlength{\fboxsep}{0pt}\colorbox{white!0}{\parbox{0.45\textwidth}{
\colorbox{red!0}{\strut A} \colorbox{red!0}{\strut fun} \colorbox{red!0}{\strut family} \colorbox{red!0}{\strut movie} \colorbox{red!0}{\strut that's} \colorbox{red!0}{\strut suitable} \colorbox{red!0}{\strut for} \colorbox{red!0}{\strut all} \colorbox{red!0}{\strut ages--} \colorbox{red!0}{\strut a} \colorbox{red!0}{\strut movie} \colorbox{red!0}{\strut that} \colorbox{red!0}{\strut will} \colorbox{red!0}{\strut make} \colorbox{red!0}{\strut you} 
\colorbox{red!0}{\strut laugh,} 
\colorbox{red!0}{\strut cry} 
\colorbox{red!0}{\strut and} 
\colorbox{red!0}{\strut realize,} 
\colorbox{red!0}{\strut `it's} 
\colorbox{red!0}{\strut never} 
\colorbox{red!0}{\strut too} 
\colorbox{red!0}{\strut late} 
\colorbox{red!0}{\strut to} 
\colorbox{red!0}{\strut believe} 
\colorbox{red!0}{\strut in} 
\colorbox{red!0}{\strut your} 
\colorbox{red!0}{\strut dreams.'} 
}}}\end{CJK*} 
& 0.00 & 0.00 \\ 
\begin{CJK*}{UTF8}{gbsn}
{\setlength{\fboxsep}{0pt}\colorbox{white!0}{\parbox{0.45\textwidth}{
\colorbox{red!0}{\strut It} \colorbox{red!0}{\strut is}
\colorbox{red!0}{\strut that}
\colorbox{red!0}{\strut rare}
\colorbox{red!0}{\strut combination}
\colorbox{red!0}{\strut of}
\colorbox{red!0}{\strut bad}
\colorbox{red!0}{\strut writing,}
\colorbox{red!0}{\strut bad}
\colorbox{red!0}{\strut direction}
\colorbox{red!0}{\strut and}
\colorbox{red!0}{\strut bad}
\colorbox{red!0}{\strut acting}
\colorbox{red!0}{\strut --}
\colorbox{red!0}{\strut the}
\colorbox{red!0}{\strut trifecta}
\colorbox{red!0}{\strut of}
\colorbox{red!0}{\strut badness.}
}}}\end{CJK*} 
& 0.00 & 0.00 &
\begin{CJK*}{UTF8}{gbsn}
{\setlength{\fboxsep}{0pt}\colorbox{white!0}{\parbox{0.45\textwidth}{
\colorbox{red!0}{\strut It's} \colorbox{red!0}{\strut a} \colorbox{red!0}{\strut cool} \colorbox{red!0}{\strut event} \colorbox{red!0}{\strut for} \colorbox{red!0}{\strut the} \colorbox{red!0}{\strut whole} \colorbox{red!0}{\strut family.} 
}}}\end{CJK*}
& 0.00 & 0.00 \\ 
\begin{CJK*}{UTF8}{gbsn}
{\setlength{\fboxsep}{0pt}\colorbox{white!0}{\parbox{0.45\textwidth}{
\colorbox{red!0}{\strut This} \colorbox{red!0}{\strut thing} \colorbox{red!0}{\strut is} \colorbox{red!0}{\strut virtually} \colorbox{red!0}{\strut unwatchable.} 
}}}\end{CJK*}
& 0.00 & 0.00 &
\begin{CJK*}{UTF8}{gbsn}
{\setlength{\fboxsep}{0pt}\colorbox{white!0}{\parbox{0.45\textwidth}{
\colorbox{red!0}{\strut Good} \colorbox{red!0}{\strut fun,} \colorbox{red!0}{\strut good} \colorbox{red!0}{\strut action,} \colorbox{red!0}{\strut good} \colorbox{red!0}{\strut acting,} \colorbox{red!0}{\strut good} \colorbox{red!0}{\strut dialogue,} \colorbox{red!0}{\strut good} \colorbox{red!0}{\strut pace,} \colorbox{red!0}{\strut good} \colorbox{red!0}{\strut cinematography.} 
}}}\end{CJK*} 
& 0.00 & 0.00 \\
\bottomrule

\end{tabular}
}
\end{table*}

\vspace{-0.2cm}
\section{Related Work}
\vspace{-0.2cm}



\textbf{Retrieval-enhanced PLMs.}\quad
Our pipeline is partly inspired by discrete demonstration methods such as ~\cite{gao2020making,DBLP:journals/corr/abs-2101-06804,kumar-talukdar-2021-reordering,le2021demoner} that retrieves few training examples in a natural language prompt, while we propose neural demonstration for enhancing the input to alleviate the limitations of input length.
Another line researches of retrieval augmentation~\cite{DBLP:journals/corr/abs-2002-08909,DBLP:conf/emnlp/KarpukhinOMLWEC20,DBLP:conf/nips/LewisPPPKGKLYR020,DBLP:journals/corr/abs-2112-08633,RETRO} retrieve useful information from a external knowledge corpus (e.g., Wikipedia) for a particular task (e.g., an open-domain question). Unlike these works,  we focus on retrieving examples from the internal training data.
Besides, semi-parametric methods~\cite{DBLP:conf/iclr/KhandelwalLJZL20,he2021efficient,DBLP:conf/iclr/KhandelwalFJZL21,DBLP:conf/emnlp/KassnerS20,alon2022neurosymbolic,meng2021gnnlm} have risen to leverage $k$-nearest neighbor classifier, a classic non-parametric algorithm that makes the prediction based on representation similarities, to enhance pre-trained language models in various tasks
However, unlike these models using nearest neighbors only for augmenting the process of prediction, we aim to develop a comprehensive retrieval mechanism for input, training and test process.

\textbf{Prompt learning for PLMs.}\quad
With the birth of GPT-3~\cite{DBLP:conf/nips/BrownMRSKDNSSAA20}, 
prompt learning~\cite{liu2021pre} has recently arisen to fill the gap between masked LM objective of PLMs and downstream fine-tuning objective. 
Prompt learning has achieves very impressive performance on various tasks, such as text classification~\cite{schick2020automatically,shin2020eliciting},  named entity recognition \cite{lightner,ma21template},  relation extraction \cite{ptr,chen21knowprompt}, event extraction \cite{hsu2021event,ye2021learning}, machine translation \cite{DBLP:journals/corr/abs-2110-06609} and language generation \cite{schick2020few}, especially under the setting of few-shot learning.
Moreover, continuous prompts have also been proposed~\cite{li2021prefix,lester2021power,liu2021gpt} to reduce prompt engineering, which directly appends a series of learnable continuous embeddings as prompts into the input sequence.
Our work is orthogonal to previous prompt learning approaches, which aim to optimize prompts, while we focus on the systematic study of retrieving related examples from training data to enhance prompt learning.



\section{Conclusion and Future Work}

We propose {\ours} that decouples knowledge from memorization by introducing retrieval augmentation to further improve the generalization ability of prompt learning on the input side and the whole process of model training and prediction.
{\ours}, is a straightforward yet effective retrieval method that combines both neural demonstrations, \knn{} guider for training and prediction.
Our extensive results show that it outperforms other demonstration-enhanced prompt methods and knowledge-enhanced prompt methods in few-shot, zero-shot and fully-supervised settings. 
Analyzing the essence of memorization validates the effectiveness of decoupling knowledge from memorization.
Interesting future directions include: 
1) apply to other tasks, such as QA and NLG, 
2) explore the noise data mining for unsupervised learning,
3) further improve the retrieve efficiency for large datasets, etc.


\section*{Acknowledgments}

We  want to express gratitude to the anonymous reviewers for their kind comments. 
This work was supported by National Natural Science Foundation of China (No.62206246, 91846204 and U19B2027), Zhejiang Provincial Natural Science Foundation of China (No. LGG22F030011), Ningbo Natural Science Foundation (2021J190), and Yongjiang Talent Introduction Programme (2021A-156-G). 
Our work was supported by Information Technology Center and State Key Lab of CAD\&CG, ZheJiang University.


\bibliography{reference}
\bibliographystyle{plain}



\section*{Checklist}

\begin{enumerate}

\item For all authors...
\begin{enumerate}
  \item Do the main claims made in the abstract and introduction accurately reflect the paper's contributions and scope?
    \answerYes{}
  \item Did you describe the limitations of your work?
    \answerYes{See Appendix.} 
  \item Did you discuss any potential negative societal impacts of your work?
    \answerYes{See Appendix.}
  \item Have you read the ethics review guidelines and ensured that your paper conforms to them?
    \answerYes{}
\end{enumerate}

\item If you are including theoretical results...
\begin{enumerate}
  \item Did you state the full set of assumptions of all theoretical results?
    \answerNA{}
        \item Did you include complete proofs of all theoretical results?
    \answerNA{}
\end{enumerate}

\item If you ran experiments...
\begin{enumerate}
  \item Did you include the code, data, and instructions needed to reproduce the main experimental results (either in the supplemental material or as a URL)?
    \answerYes{We include the source code and data in our supplemental material submission, and we outline the data generation procedure, the evaluation protocol, the training regime, and everything else necessary for reproduction either in the main body of the paper or in the appendix.} 
  \item Did you specify all the training details (e.g., data splits, hyperparameters, how they were chosen)? 
  \answerYes{See Subsection~\ref{subsec:details} and Appendix.} 
        \item Did you report error bars (e.g., with respect to the random seed after running experiments multiple times)?
    \answerYes{We list the standard deviation for few-shot setting. } 
        \item Did you include the total amount of compute and the type of resources used (e.g., type of GPUs, internal cluster, or cloud provider)? 
        \answerYes{We introduce type of resources in Section~\ref{subsec:details}.} 
  
\end{enumerate}

\item If you are using existing assets (e.g., code, data, models) or curating/releasing new assets...
\begin{enumerate}
  \item If your work uses existing assets, did you cite the creators?
    \answerYes{}
  \item Did you mention the license of the assets?
    \answerNo{The code and the data are proprietary.}
  \item Did you include any new assets either in the supplemental material or as a URL?
    \answerNo{}
  \item Did you discuss whether and how consent was obtained from people whose data you're using/curating?
    \answerNo{The code and the data are proprietary.}
  \item Did you discuss whether the data you are using/curating contains personally identifiable information or offensive content?
    \answerNo{}
\end{enumerate}

\item If you used crowdsourcing or conducted research with human subjects...
\begin{enumerate}
  \item Did you include the full text of instructions given to participants and screenshots, if applicable?
    \answerNA{}
  \item Did you describe any potential participant risks, with links to Institutional Review Board (IRB) approvals, if applicable?
    \answerNA{}
  \item Did you include the estimated hourly wage paid to participants and the total amount spent on participant compensation?
    \answerNA{}
\end{enumerate}

\end{enumerate}


 \appendix
 
\section{Datasets and Templates}
\label{App:dataset-template}
In this section, we introduce the datasets as shown in Table~\ref{tab:dataset_stat} and list the templates we use in experiments as follows.

\begin{table}[!htbp]
\vspace{-3mm}
\caption{Detailed dataset statistics.}
    \centering
    \scalebox{1.0}{
    \begin{tabular}{cccr}
\toprule
    Dataset &Type & \# Class & Test Size \\ 
\midrule
      SST-2   & Sentiment  & 2 & 872 \\
      MR   & Sentiment  & 2 & 2,000 \\
      CR & Sentiment  & 2 & 2,000 \\
      MNLI & NLI  & 3  & 9,815 \\
      QNLI & NLI  & 2  & 5,463 \\
      QQP  & Paraphrase  & 2  & 40,431  \\
      FewNERD  & Entity Typing  & 66 & 96,901 \\
      SemEval  & Relation Extraction  & 19 & 2,717 \\
      TACRED  & Relation Extraction  & 42 & 15,509 \\
     \bottomrule
    \end{tabular}
    }
    \label{tab:dataset_stat}
\end{table}

\textbf{SST-2, MR, CR}.\quad  For the single sentence classification tasks, we follow the LM-BFF~\cite{gao2020making} to design the templates:
\newtcolorbox{mybox}{colback=blue!8!white, colframe=orange!75!black,height=0.02\textheight}

\newtcolorbox{mybox2}{colback=gray!6!white, colframe=gray!75!black, width=0.53\textwidth}
\newtcolorbox{mybox3}{colback=gray!6!white, colframe=gray!75!black, width=0.50\textwidth}

\begin{mybox}
\vspace{-6mm}
\small
\begin{align*}
     T(\mathbf{x})  &=\text{\texttt{[CLS]}}  \mathbf{x} \ \text{ It was \texttt{[MASK]}}.    \\
\end{align*}
\end{mybox}

We set Verbalizer: (great/terrible) $\rightarrow$ (positive/negative) for SST-2 MR and CR. For the Yahoo dataset, we assign the Verbalizer following the original labels.

\textbf{MNLI, QNLI, QQP}. \quad For the sentence pair classification tasks, we follow LM-BFF~\cite{gao2020making}  to set
Verbalizer: (Yes/Maybe/No) $\rightarrow$ (entailment/neutral/contradiction), and define the following templates:
\begin{mybox}
\vspace{-6mm}
\small
\begin{align*}
     T(\mathbf{x_1},\mathbf{x_2})  &=\text{\texttt{[CLS]}}  \mathbf{x_1} ? \texttt{[MASK]}, \mathbf{x_2}    \\
\end{align*}
\end{mybox}

 \textbf{FewNERD, SemEval, TACRED}. \quad FewNERD, SemEval and TACRED  are  datasets for information extraction, which require inserting the entity into the template.  Therefore, we follow ~\cite{ding2021prompt} and ~\cite{chen21knowprompt} to define the template and verbalizers.


\section{Compared Baselines}
\label{App:baselines}

In this subsection,
we introduce the baselines we compare with and re-produce them under the same settings with their open-source codes.

LM-BFF uses several other tricks, such as prompt ensemble, while KPT utilizes tremendous external knowledge. 
We do not use any of these tricks and external knowledge since we get the most out of the data to decouple part of knowledge from parametric memorization.
Our {\ours} mechanism is orthogonal to other methodological improvements of prompt-tuning (such as continuous prompt in P-tuning~\cite{liu2021gpt} and DART~\cite{zhang2021dart} ) and can be combined with other prompt-tuning methods in future work.

\textbf{Fine-tuning (FT).}\quad The traditional fine-tuning method regard the hidden embedding of \texttt{[CLS]} token of the PLM  as the representation of the sentence and then feeds them into a classification layer to make predictions.

\textbf{LM-BFF.}\quad LM-BFF~\cite{gao2020making} is a typical prompt-tuning method wrapping an input sentence into a handcrafted template. 
Here we re-produce LM-BFF based on their open-source codes~\footnote{\url{https://github.com/princeton-nlp/LM-BFF}}  with the same manual prompts as {\ours}  for a fair comparison.

\textbf{LM-BFF (+Demo).}\quad
This approach is the  above LM-BFF \cite{gao2020making}  combined with the  demonstration~\cite{DBLP:conf/nips/BrownMRSKDNSSAA20}.
Different from {\ours}, it uses examples of natural language as demonstrations, which is restricted by the input length of the language model. 
Thus, LM-BFF (+demo) is not suitable for multi-class classification tasks. 

\textbf{KnowPrompt.}\quad KnowPrompt~\cite{chen21knowprompt} is a SOTA prompt-tuning method for relation extraction tasks with multiple classes.
We apply our {\ours} over KnowPrompt on information extraction tasks for comparison, aiming to verify the broad applicability of our method.

\textbf{Incorporating Knowledge into Prompt  (KPT).} 
KPT~\cite{KPT} focuses on incorporating external knowledge into the verbalizer by refining the expanded label word space to improve and stabilize prompt-tuning, which is a solid baseline for comparison. 
We follow their public codes\footnote{\url{https://github.com/ShengdingHu/KnowledgeablePromptTuning}} to conduct experiments in the same setting for a fair comparison.

 \textbf{LOTClass.}\quad LOTClass~\cite{meng2020text} is the SOTA method in unsupervised text classification that utilizes the PLM to extract the label-related words from the whole unlabeled training corpus. 
 Then it leverages the Masked Category Prediction task to \textbf{train} on the unlabeled corpus with pseudo labels.

\section{Experimental Settings}
\label{app:exp_settings}
We report the hyper-parameters   in Table~\ref{tab:app_exp_settings}. Most of the hyper-parameters are the default parameters of LM-BFF\footnote{\url{https://github.com/princeton-nlp/LM-BFF}}.

\begin{table}[!htbp]
\vspace{-3mm}
\caption{Hyper-parameter settings.}
    \centering
    \begin{tabular}{c|c}
\toprule
      Hyper-parameter   & Value \\
\midrule
   maximum sequence length      &  \{128, 256\} \\
   max training step & 800 \\
   evaluation step & 80 \\
   learning rate & \{1e-5, 2e-5, 5e-5\} \\
   batch size & \{2, 4, 8\} \\
   adam epsilon & 1e-8 \\
\bottomrule
    \end{tabular}
    \label{tab:app_exp_settings}
        \vspace{-3mm}
\end{table}

\section{Tuning Retrieve Parameters}
\label{App:hyper}
The final distribution of the label is affected by the hyperparameters of $\beta$, $k$ and $\lambda$ when conducting \knn{}-train and \knn{}-test. 
Thus, we provide insight into the effect of $\beta$, $k$ and $\lambda$ on the final results.

\begin{figure*}[!t]
    \centering
    \subfigure[{$\beta$} varies.]{
    \includegraphics[width=0.25\textwidth]{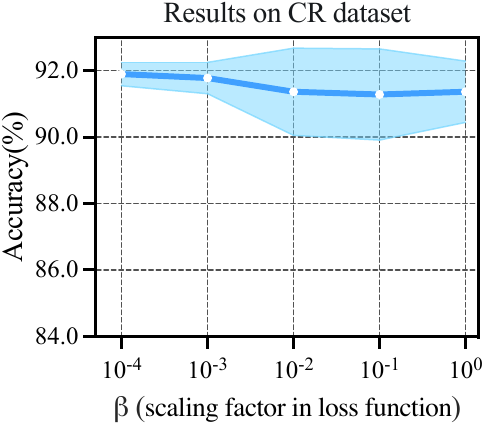}
    \label{fig:analysis_beta}
    }
    \hspace{-15pt}
    \quad
    \subfigure[{$\lambda$} varies.]{
    \includegraphics[width=0.25\textwidth]{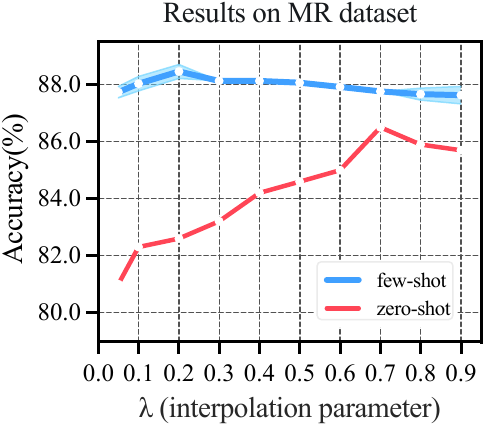}
    \label{fig:analysis_lambda}
    }
    \hspace{-15pt}
    \quad
    \subfigure[{$k$} varies.]{
     \includegraphics[width=0.25\textwidth]{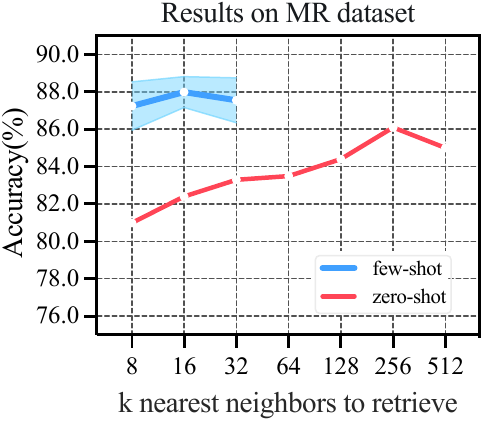}
     \label{fig:analysis_topk}
     }
     \caption{\label{fig:analysis} Effect of the hyperparameters of the retrieval.
    }
\end{figure*}

\textbf{$\beta$ varies:}
Figure \ref{fig:analysis_beta} shows the performance of the model when the $\beta$ increases and reveals that the model
performs worse as the  $\beta$ increases
on a 16-shot CR dataset. This finding indicates that a moderate degree of \knn{} guiding training is essential since \knn{} can help the model attend to hard examples, but excessive attendance of \knn{}-train also can bring the noise.

\textbf{$\lambda$ varies:}
From Figure \ref{fig:analysis_lambda}, we observe that model achieves optimal results on a 16-shot MR dataset when $\lambda$ is set to be  0.2 while attaining the best results on MR in the zero-shot setting when $\lambda$ is set to be  0.7.
We think the model may require more reference when there is no data for training.

\textbf{$k$ varies:}
As shown in  Figure~\ref{fig:analysis_topk}, the model performance in the 16-shot MR dataset fluctuates very little. 
In contrast, the result in the zero-shot MR dataset continues to improve as $k$ increases until it converges when reaching a threshold ($k=256$).
It illustrates that the $k$-NN retrieval provides more evidence for reference in zero-setting.

{\color{blue}\section{ Discussion of Limitation}}

\paragraph{Analysis of Efficiency}
\label{anlysis:efficiency}
 We make the comparison between LM-BFF (man), LM-BFF (+demo) and {\ours} in speed on the MR dataset for the 16-shot setting.
We observe that the speed of {\ours} and LM-BFF (+demo) are approximately 1.12 and 20 times slower than LM-BFF (man) on the 16-shot MR dataset.
The slow inference of LM-BFF (+demo) is due to the fact that they sample from the top $r\%$ instances ($r = 50$) for each class to use as demonstrations and vastly increase the length of the input, thus, increasing computational complexity significantly.
And the bottleneck of computational speed is general limitations of retrieval methods, and our method is no exception.
We will leave the engineering optimization about retrieval speed in our future work.

{\color{blue}\paragraph{Analysis of memory usage}
Actually, our method adopt FAISS tools for retrieval. FAISS  is an excellent open-source library for fast nearest neighbor retrieval in high-dimensional spaces, which supports searching only from RAM, which involves k-means clustering for improving memory usage efficiency. Memory usage is negligible in the few-shot settings and acceptable in the full-data settings. Our retrieval process is performed mainly on CPU, and we compare the utilization of CPU with and without retrieval in the SST-2 full setting as follows:
\begin{itemize}

\item  The CPU utilization was 46.2\% with the retrieval process and 2.5\% without it (Our CPU is Intel(R) Xeon(R) Silver 4210R CPU @ 2.40GHz with 40 cores).

\item In terms of memory usage, adding retrieval requires about 2.5G more memory than not. One way to reduce resource usage is to store the datastore on the disk in advance, then read and release it in the retrieval process.
\end{itemize}
}




\end{document}